\documentclass[letterpaper, 10 pt, conference]{ieeeconf}  

\IEEEoverridecommandlockouts                              

\usepackage{cite}
\usepackage{amsmath,amssymb,amsfonts}
\usepackage{algorithmic}
\usepackage{graphicx}
\newcommand{\NA}{N/A}
\usepackage{textcomp}
\usepackage{xcolor}
\usepackage{colortbl}
\usepackage{verbatim}
\usepackage{booktabs}
\usepackage{wrapfig}
\usepackage{url}
\usepackage{makecell}

\title{\LARGE \bf
FoldBack: Self-Correcting Masked Generative Policy for Long-Horizon Garment Folding
}

\author{
Lipeng Zhuang,
Shiyu Fan,
Yingdong Ru,
Zhuo He,\\
Florent P. Audonnet,
Paul Henderson,
Gerardo Aragon Camarasa%
\thanks{The authors are with the University of Glasgow, Glasgow, United Kingdom.
\texttt{2658047z@student.gla.ac.uk}}
}

\newcommand{\lp}[1]{\textcolor{red}{#1}}

\begin{document}

\maketitle
\thispagestyle{empty}
\pagestyle{empty}



\begin{abstract}
We present FoldBack, a self-correcting masked generative policy for long-horizon garment folding. Existing long-trajectory policies may continue after a missed or slipped grasp even when the garment has not reached the intended configuration. We structure FoldBack's recovery mechanisms around three inference-time decisions: \emph{when to refine and verify}, \emph{how to roll back}, and \emph{where and how to retry}. FoldBack aligns refinement and grasp verification with pick-and-place events, returns the robot to a retryable pre-grasp configuration while preserving successful grasps, and selectively regenerates the failed segment and selected future actions while avoiding previous failed grasp locations. 
To our knowledge, FoldBack is the first editable full-trajectory policy to unify these decisions, enabling failed interactions to be detected, undone, and repaired before execution continues, without recovery demonstrations or base-policy retraining.
Across 33 real garments from six categories, FoldBack achieves \(75.2\%\) final folding success and \(0.837\) final-mask IoU, versus \(45.7\%\) and \(0.689\) for the strongest prior baseline. 
\end{abstract}

\begin{figure*}[t]
    \centering
    \includegraphics[width=0.96\linewidth]{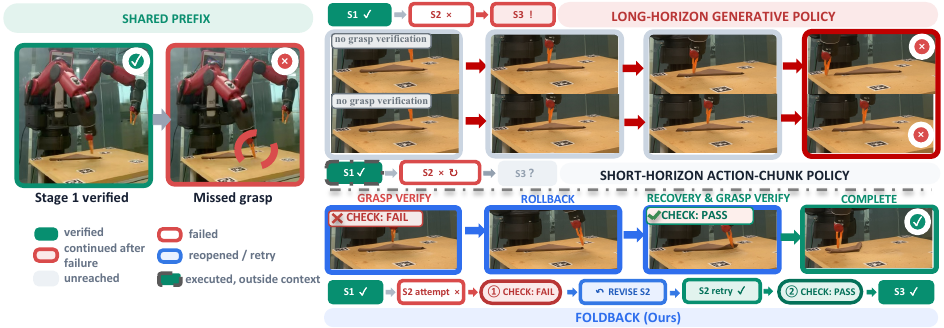}
    
    \vspace{-6pt}
\caption{
Failure handling in garment folding.
Unlike baseline policies that may continue or stall after a failed grasp, FoldBack verifies the grasp outcome, rolls back, repairs the trajectory, and retries before advancing.}
    \label{fig:teaser}
    \vspace{-5mm}
\end{figure*}

\section{Introduction}
Garment folding is a long-horizon manipulation task that requires executing a sequence of precise single-arm or bimanual pick-and-place actions. Each action changes the garment configuration, which in turn determines the conditions for subsequent folding actions. A missed or slipped grasp therefore invalidates subsequent folding actions \cite{DBLP:journals/ral/ChenXW26}.
For example, a robot may close its gripper and lift without successfully moving the garment, leaving it in a configuration incompatible with the next folding action. 
Reliable folding not only requires long-horizon action generation but also a mechanism for detecting grasp failures and recovering from them before execution advances. 

Visuomotor imitation learning enables robots to learn complex behaviors from demonstrations~\cite{DBLP:conf/corl/XueLX0ZL23,DBLP:journals/ral/ChenXW26,DBLP:conf/icra/ZhuangFRAHA25}. Short-horizon policies generate action chunks myopically given recent observations, enabling reactive execution~\cite{DBLP:journals/corr/abs-2501-14400,DBLP:conf/rss/ZhaoKLF23,DBLP:journals/ijrr/ChiXFCDBTS25,DBLP:conf/rss/ZeZZHWX24}. However, grasp failures can produce out-of-distribution observations that cause subsequent predictions to stall, drift, or lose coherence with the overall folding strategy. Long-horizon generative policies generate complete, coherent action sequences and can refine future action tokens as observations arrive~\cite{DBLP:conf/corl/XianGGKF23,DBLP:journals/corr/abs-2512-09101}; yet executed action tokens remain fixed in the trajectory history without verifying the intended manipulation outcome. Thus, execution may advance to the next folding stage without revisiting the failed interaction. Runtime monitors can detect such failures but do not themselves revise the policy trajectory~\cite{agia2024unpacking,xu2025can,DBLP:conf/icra/GokmenHK23}, while recovery-aware methods often rely on additional recovery data, policy adaptation, or large-model reasoning~\cite{DBLP:journals/ral/ChenXW26,dai2025racer,liu2023reflect,ma2026cyclevla,hao2026far}.

Motivated by these limitations, we introduce FoldBack, an outcome-gated masked generative policy for long-horizon garment folding that supports recovery from grasp failures. FoldBack represents the full action trajectory as discrete tokens and maintains a grasp-verified history, a tentative grasp-critical window, and an editable future. Executing a grasp-critical window does not immediately add it to history. The window is added only after all required grasps are verified; otherwise, the history remains unchanged and the failed window stays editable (Fig.~\ref{fig:teaser}). 

FoldBack is then 
organized around three inference-time decisions which constitute the contributions of this paper:
\begin{enumerate}
    \item \textit{when} to refine and verify (Sec.~\ref{sec:key_action}),
    \item \textit{how} to roll back after a failure (Sec.~\ref{sec:rollback}), and 
    \item \textit{where and how} to retry (Sec.~\ref{sec:plan_repair}).
\end{enumerate}
This enables recovery without recovery demonstrations or base-policy retraining. Indeed, across 33 garments from 6 categories, FoldBack achieves \(75.2\%\) final folding success and a mean final-mask IoU of \(0.837\) across single-arm and bimanual folding tasks, outperforming the strongest prior baseline by \(29.5\) percentage points in final folding success and \(0.148\) in mean final-mask IoU. 
FoldBack also generalizes to unseen garment instances, initial orientations, and non-canonical configurations as shown in Sec.~\ref{sec:experiments}.

\section{Related Work}

\noindent\textbf{Learning-based garment manipulation.}
Garment-folding policies typically compose prescribed primitives or generate local action segments \cite{DBLP:journals/ral/MoXWDGL23,DBLP:conf/icra/CanberkCHBCFS23,DBLP:conf/corl/WengBWAH21}. SpeedFolding~\cite{DBLP:conf/iros/AvigalBAKG22} and UniFolding~\cite{DBLP:conf/corl/XueLX0ZL23} execute parameterized primitives, while BiFold~\cite{DBLP:conf/icra/BarbanyCT25} predicts a language-conditioned pick-and-place action at each step. Model-based methods learn cloth dynamics for planning~\cite{DBLP:conf/corl/LinWHH21,DBLP:conf/rss/HuangLH22}, while UniClothDiff~\cite{DBLP:journals/corr/abs-2503-11999} uses diffusion models for generative state estimation and dynamics prediction within model-predictive control. Other generative policies model distributions over action primitives (DeformPAM~\cite{DBLP:conf/icra/ChenXZFL25}), closed-loop or local manipulation trajectories (FoldNet~\cite{DBLP:journals/ral/ChenXW26} and HALO~\cite{DBLP:conf/nips/WangWCWLZGMAD25}), or stage-wise point-cloud trajectories (MetaFold~\cite{DBLP:conf/iros/ChenLWLHXGGWXHS25}).
These methods are effective for local or stage-wise manipulation, but long-horizon folding requires consistent execution across multiple strategies with both single- and bimanual actions. 
FoldBack addresses this with full-trajectory masked generation for an editable plan (Sec.~\ref{sec:method}), Key-Action-Aligned Inference for event-aligned refinement and grasp verification (Sec.~\ref{sec:key_action}), Verified Rollback to keep failed windows revisable (Sec.~\ref{sec:rollback}), and Structured Plan Repair to selectively revise the affected trajectory (Sec.~\ref{sec:plan_repair}).

\noindent\textbf{Imitation learning.}
Imitation learning learns visuomotor policies from demonstrations. Short-horizon policies generate local action sequences and replan from observations, including ACT~\cite{DBLP:conf/rss/ZhaoKLF23}, Diffusion Policy~\cite{DBLP:journals/ijrr/ChiXFCDBTS25}, DP3~\cite{DBLP:conf/rss/ZeZZHWX24}, 3D Diffuser Actor~\cite{ke2024diffuseractor}, BeT~\cite{shafiullah2022bet}, and VQ-BeT~\cite{lee2024vqbet}. 
Recent methods improve temporal consistency or inference efficiency through sequence-level generation and action-chunk refinement~\cite{DBLP:conf/corl/XianGGKF23,DBLP:conf/rss/PrasadLWZB24,zhang2025flowpolicy,black2025training,DBLP:conf/nips/BlackGL25}, but do not maintain an editable task plan.
MGP-Long~\cite{DBLP:journals/corr/abs-2512-09101} instead generates a complete trajectory, fixes executed tokens as history, and refines uncertain future tokens. Thus, it cannot revisit failed grasps. FoldBack addresses this by adding grasp-critical windows to the grasp-verified history only after verification (Sec.~\ref{sec:rollback}) and regenerating failed windows during plan repair (Sec.~\ref{sec:plan_repair}).

\noindent\textbf{Failure monitoring and recovery.}
Runtime monitoring methods detect anomalous behavior or insufficient progress but do not generate recovery actions \cite{agia2024unpacking,xu2025can,DBLP:conf/icra/GokmenHK23}. Some recovery-aware methods instead learn corrective behavior from additional data. For instance, FoldNet uses KG-DAgger recovery trajectories~\cite{DBLP:journals/ral/ChenXW26}, while RACER~\cite{dai2025racer} trains a VLM-supervised, language-conditioned policy on generated recovery trajectories and fine-grained annotations. More general correction frameworks include CycleVLA~\cite{ma2026cyclevla}, which combines VLA-based progress prediction with VLM-guided backtracking, while FAR~\cite{hao2026far} adapts a diffusion policy from test-time failures and perturbs retry actions. Bellman-Guided Retrials~\cite{DBLP:journals/corr/abs-2406-15917} performs progress-guided retrials, while Rewind-IL~\cite{DBLP:journals/corr/abs-2604-16683} rewinds policies to verified checkpoints.
As described in Sec.~\ref{sec:rollback} and Sec.~\ref{sec:plan_repair}, FoldBack instead reopens failed windows within an editable full trajectory and selectively repairs the affected trajectory while preserving grasp-verified history.
It requires neither failure-recovery demonstrations, base-policy adaptation, nor auxiliary LLM/VLM reasoning.

\begin{figure*}[t]
    \centering
    \includegraphics[width=0.95\linewidth]{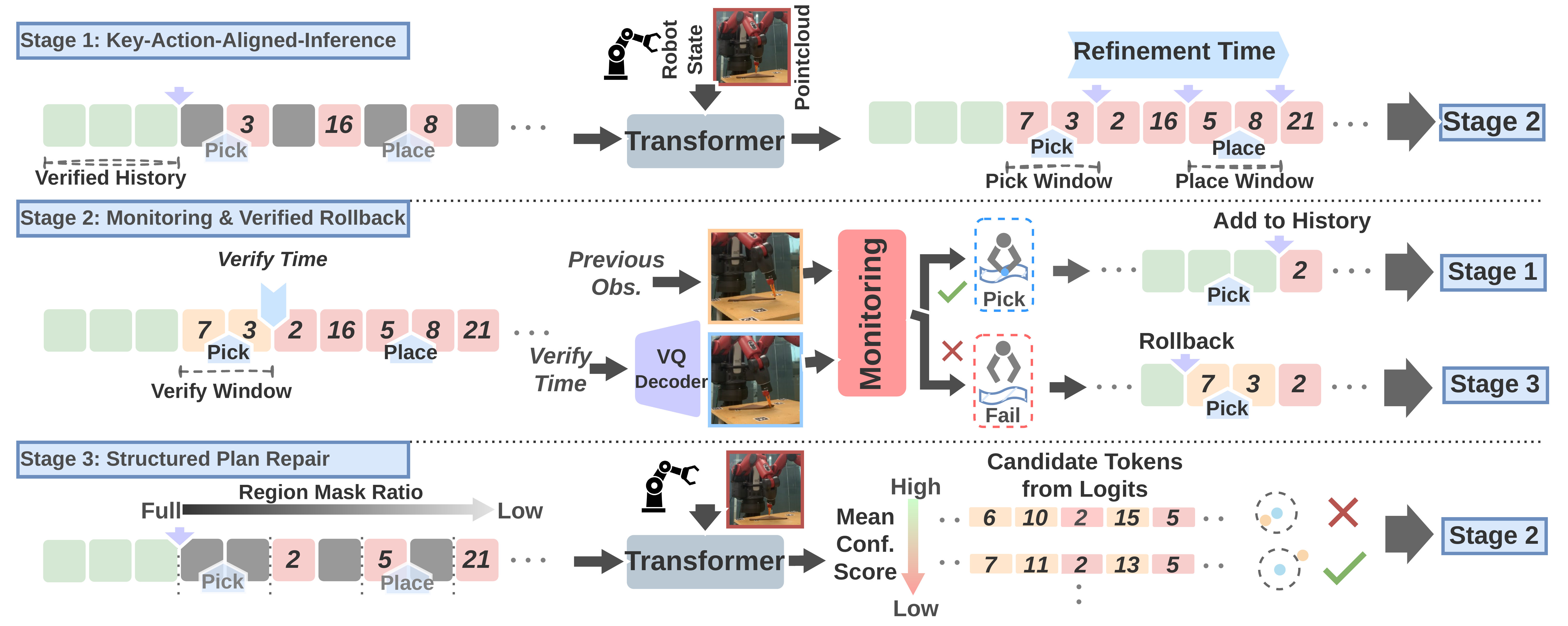} 
    \vspace{-6pt}
    \caption{
Overview of FoldBack. Key-Action-Aligned Inference aligns trajectory refinement with pick-and-place events; grasp monitoring and verified rollback determine whether a grasp-critical window is added to the verified trajectory history or kept editable for recovery; structured plan repair then remasks and resamples the editable trajectory after failure.
}
    \label{fig:method}
    \vspace{-5mm}
\end{figure*}

\section{Method}
\label{sec:method}

We formulate garment folding as long-horizon action generation conditioned on an observation \(\mathbf{o}_t=(\mathbf{P}_t,\mathbf{s}_t)\) at time \(t\), where \(\mathbf{P}_t\) is an observed point cloud and \(\mathbf{s}_t\) is the
robot's proprioceptive state. An action \(\mathbf{a}_t\in\mathbb{R}^{16}\) consists of, for each arm \(r\in\{L,R\}\), a 3-D end-effector position, a 4-D orientation quaternion, and a binary gripper state \(g_t^r\in\{0,1\}\).
From human demonstrations, we learn a policy that predicts a \(T\)-step action trajectory \((\mathbf{a}_t,\ldots,\mathbf{a}_{t+T-1})\) conditioned on the current observation \(\mathbf{o}_t\).


FoldBack adopts the Masked Generative Policy (MGP)~\cite{DBLP:journals/corr/abs-2512-09101} as its generative backbone.
MGP represents a \(T\)-step action trajectory using \(N\) discrete tokens
\(\mathbf{y}=(y_1,\ldots,y_N)\), with a VQ-VAE tokenizer~\cite{van2017neural}
trained to map between continuous action trajectories and this discrete representation, and \(N<T\). 
Given a partially masked token sequence and observation \(o_t\), the Masked Generative Transformer (MGT) predicts masked tokens in parallel. MGP initially generates the full trajectory from a fully masked sequence, executes decoded actions, and subsequently remasks selected future tokens using updated observations.

In the original MGP inference procedure, executed action tokens become fixed trajectory history regardless of whether their intended physical outcomes were achieved. FoldBack instead keeps a grasp-critical window tentative until its grasp outcome is verified. If all required grasps succeed, the window is added to the grasp-verified trajectory history; otherwise, the history remains unchanged and the failed window remains editable for recovery. As shown in Fig.~\ref{fig:method}, FoldBack determines when to refine and verify the trajectory (Key-Action-Aligned Inference, Sec.~\ref{sec:key_action}), whether to add the executed window to history or return the robot to the pre-grasp configuration (Sec.~\ref{sec:rollback}), and how to repair the editable trajectory after failure (Sec.~\ref{sec:plan_repair}).

We train the MGP backbone on Flat'n'Fold~\cite{DBLP:conf/icra/ZhuangFRAHA25} following its original two-stage procedure~\cite{DBLP:journals/corr/abs-2512-09101}. The VQ-VAE tokenizer is first trained with position and orientation reconstruction losses, binary cross-entropy on gripper states, and the VQ commitment loss, after which it is frozen. 
The MGT is then 
trained to recover action tokens that are randomly masked or replaced with alternative codebook indices,
conditioned on the corresponding observation. 
FoldBack requires no additional policy training or recovery demonstrations.

\subsection{Key-Action-Aligned Inference (KAAI)}
\label{sec:key_action}

\emph{KAAI} ensures trajectory refinement and grasp verification happen after pick-and-place transitions of garment folding rather than at uniform execution intervals. Since pick and place events mark the main physical transitions in folding, after which the garment configuration can change substantially, this effectively enables the robot to perform visual servoing at key moments in the trajectory.
Following initial trajectory generation, FoldBack extracts these key actions from the decoded gripper-state sequences \((g_1^r,\ldots,g_T^r)\). An open-to-closed transition of the gripper defines a \textsc{Pick}, whereas a closed-to-open transition defines a \textsc{Place}. This yields \(J\) key actions at times \(t_1<\cdots<t_J\). 

Each key action \(k\) defines an action window \(\mathcal{W}_k=\{t_k-4,\ldots,t_k+3\}\). Let \(\mathcal{I}_k\subseteq\{1,\ldots,N\}\) denote the smallest set of token indices whose decoded actions cover \(\mathcal{W}_k\). In bimanual manipulation, the two arm-specific \textsc{Pick} transitions are treated as a single key action, whose action window spans from the first gripper-closing event to the end of the corresponding joint lift. 
A window whose key action is a \textsc{Pick} is \emph{grasp-critical} because it establishes a grasp that the rest of the folding stage depends on.
Before and after executing each key-action window \(\mathcal{W}_k\),
FoldBack refines the remaining unexecuted tokens using the latest
observation. After each refinement, FoldBack re-decodes the gripper-state sequence and recomputes the transition time \(t_j\), action window \(\mathcal{W}_j\), and token set \(\mathcal{I}_j\) for each unexecuted key action \(j\). The \textsc{Pick}/\textsc{Place} order and arm assignment remain fixed, while their timing may shift as the trajectory
is refined. We use two MGP mask-refinement iterations at each update.

\vspace{-1mm}
\subsection{Grasp-outcome monitoring and Verified Rollback}
\label{sec:rollback}

FoldBack verifies each required grasp after lifting, and then adds the corresponding grasp-critical window to the trajectory history only when all grasps succeed. In sequential pick-and-place manipulation, each grasp is verified after its own lift. In bimanual manipulation, the first arm maintains its grasp while the second performs its pick; both arms then lift, and both grasps are verified. If any grasp fails, FoldBack initiates the verified rollback procedure described below.

\begin{figure}[t]
    \centering
    \includegraphics[width=0.95\linewidth]{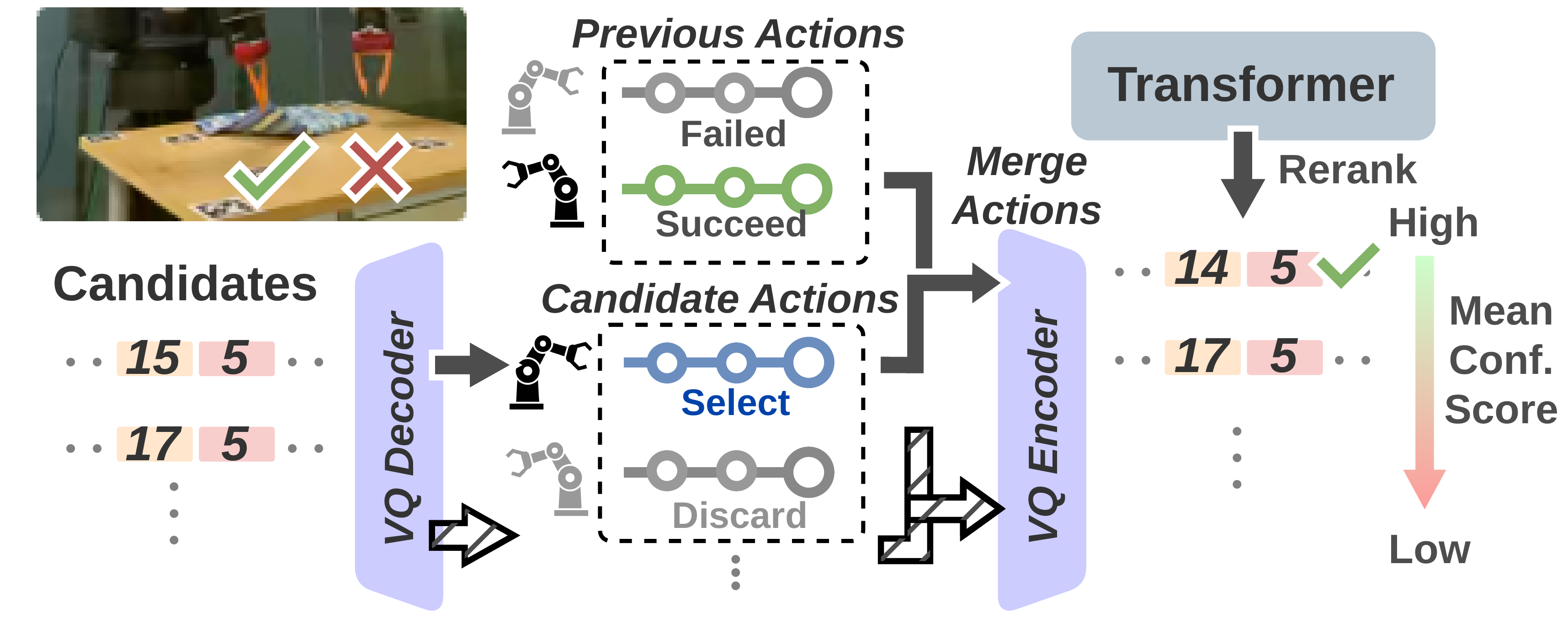}   
    \vspace{-6pt}
    \caption{Repair under an asymmetric bimanual grasp failure.
}
    \label{fig:bimanual}
    \vspace{-6mm}
\end{figure}

\paragraph{Local Grasp-Outcome Monitor}
At verification point \(k\), FoldBack evaluates each arm \(r\) whose grasp must be verified. Using the gripper poses, we extract pre-pick and post-lift local point clouds,
\(\mathbf{P}_{k,r}^{\mathrm{pre}}\) and \(\mathbf{P}_{k,r}^{\mathrm{lift}}\),
from a fixed region below the fingertips before grasp closure and after lifting, respectively. The two point clouds are expressed in their respective gripper frames. This representation focuses the monitor on changes in local garment geometry, largely excludes the gripper, and requires no garment segmentation. Thus, a PointNet-based ~\cite{qi2017pointnet} binary classifier \(f_{\psi}\) predicts
\begin{equation}
p_{k,r}
=
f_{\psi}\!\left(
\mathbf{P}_{k,r}^{\mathrm{pre}},
\mathbf{P}_{k,r}^{\mathrm{lift}}
\right)
\in[0,1],
\label{eq:grasp_monitor}
\end{equation}
where \(\psi\) denotes the monitor parameters and \(p_{k,r}\) is the predicted
grasp-success probability. The grasp is verified as successful when
\(p_{k,r}\geq\tau_{\mathrm{mon}}\), with \(\tau_{\mathrm{mon}}=0.50\) selected on the validation set and fixed for all evaluations.

We train \(f_{\psi}\) using binary cross-entropy on successful Flat'n'Fold demonstrations. A positive pair contains \(\mathbf{P}_{k,r}^{\mathrm{pre}}\) and its corresponding \(\mathbf{P}_{k,r}^{\mathrm{lift}}\). 
A negative pair replaces \(\mathbf{P}_{k,r}^{\mathrm{lift}}\) with a local point cloud from a nearby pre-pick frame at the location where the post-lift point cloud would be taken.
Because the garment remains on the table in that frame, the resulting pair approximates a missed grasp without requiring physical failure trajectories. Training pairs are constructed independently for each arm, and one classifier is shared across single-arm and bimanual stages. The trained policy and monitor remain fixed during deployment.

\paragraph{Grasp-Verified Trajectory History}
FoldBack does not immediately add the executed tokens of a grasp-critical window to the trajectory history. Instead, it keeps the token subsequence \(\mathbf{y}_{\mathcal{I}_k}\) tentative until post-lift verification is complete, while the unexecuted future trajectory remains editable. Let \(v_k\in\{0,1\}\) denote the grasp-outcome monitor decision for \(\mathcal{W}_k\), where \(v_k=1\) indicates that all associated grasps are verified as successful and \(v_k=0\) otherwise. 
If \(v_k=1\), FoldBack appends \(\mathbf{y}_{\mathcal{I}_k}\) to the grasp-verified trajectory history \(\mathcal{H}^{\mathrm{ver}}\); otherwise, \(\mathcal{H}^{\mathrm{ver}}\) remains unchanged and the failed window remains editable for plan repair; neither the failed attempt nor the rollback motion is added to the history.
Tokens executed outside the current grasp-critical window remain fixed as trajectory history.
For a bimanual manipulation window, the joint token subsequence \(\mathbf{y}_{\mathcal{I}_k}\) is saved to the trajectory history only if both arms' grasps are verified as successful.

\paragraph{Verified rollback}
FoldBack records the executed end-effector motion from the pre-grasp to the post-lift configuration. When \(v_k=0\), each failed gripper opens and the recorded motion is executed in reverse, restoring the pre-grasp configuration. In a bimanual manipulation, both arms return in coordination to their pre-grasp configurations; if only one grasp fails, the successful gripper remains closed to preserve its grasp. 
At the restored configuration, FoldBack captures an updated point cloud and robot state. 
The procedure is policy-agnostic and requires no base-policy retraining. After rollback, action-chunking policies replan from the updated observation, whereas masked long-trajectory policies can additionally retain grasp-verified history and regenerate affected editable tokens.

\subsection{Structured Trajectory Plan Repair}
\label{sec:plan_repair}

Verified rollback restores a retryable configuration but does not revise the trajectory. This is because the unexecuted actions were generated under the assumption that the failed grasp would succeed. 
When a grasp-critical window \(\mathcal{W}_k\) fails verification, FoldBack repairs the trajectory in two steps.

\paragraph{Distance-guided remasking}
The unexecuted token positions after \(\mathcal{I}_k\) are grouped into segments \(\mathcal{S}_1,\ldots,\mathcal{S}_{J-k}\), one for each subsequent key-action window. 
\(\mathcal{S}_1\) contains the actions immediately after the failed window up to the next key-action window; each subsequent segment spans between consecutive key-action windows.
The final segment additionally includes any remaining actions after the last key-action window. FoldBack fully remasks \(\mathcal{I}_k\) and assigns each segment \(\mathcal{S}_{\ell}\) the masking ratio, defined as the fraction of action tokens masked and regenerated:
\begin{equation}
\rho_{\ell}
=
\rho_{\mathrm{base}}
+
\left(\rho_{\mathrm{near}}-\rho_{\mathrm{base}}\right)
\alpha^{\ell-1},
\label{eq:distance_mask_ratio}
\end{equation}
where \(\rho_{\mathrm{base}}\) is the masking ratio used by MGP-Long during trajectory refinement, \(\rho_{\mathrm{near}}>\rho_{\mathrm{base}}\) is the ratio applied to \(\mathcal{S}_1\), and \(\alpha\in(0,1)\) controls the rate of decay. The masking ratio therefore decreases toward \(\rho_{\mathrm{base}}\) as the segment distance from the failure increases.

Within each segment \(\mathcal{S}_{\ell}\), FoldBack remasks the \(\rho_{\ell}\)-fraction of tokens with the lowest stored confidence \(\gamma_i^{\mathrm{last}}\), defined as the probability assigned to the current token \(y_i\) when it was most recently generated. We denote the resulting set of remasked positions, including \(\mathcal{I}_k\), by \(\mathcal{M}\). FoldBack regenerates the tokens at these positions using the observation captured after the robot returns to the pre-grasp configuration, while the grasp-verified prefix and unselected future tokens remain fixed as anchors. During subsequent mask-refinement iterations, the stored confidence of each regenerated token is updated with its newly predicted probability. We use \(\rho_{\mathrm{near}}=0.80\),
\(\rho_{\mathrm{base}}=0.30\), and
\(\alpha=0.50\) in all experiments.

\paragraph{Failure-aware candidate resampling}
A regenerated trajectory plan may repeat the grasp that just failed. FoldBack therefore draws eight repair candidates for \(\mathcal{M}\) from the post-rollback observation. Each arm maintains an episode-level memory \(\mathcal{F}_r\) of its failed table-plane pick locations, to which the current failure is added before sampling. A candidate is rejected if its proposed pick for an arm that just failed lies within an exclusion radius \(\epsilon=0.04\,\mathrm{m}\) of any location in \(\mathcal{F}_r\). Under an asymmetric bimanual failure, only candidates for the failed arm are rejected based on proximity to previous failed pick locations, while the successful arm retains its grasp and is not resampled.
FoldBack samples up to three candidate batches to obtain a valid retry trajectory. If no valid candidate is found, the recovery episode is marked as unsuccessful and the manipulation terminates at the restored pre-grasp configuration.

Let \(\gamma_i^{(q)}\) denote the probability assigned to candidate (\(q\in\{1,\ldots,8\}\)) at regenerated position \(i\in\mathcal{M}\). Each candidate is scored by its mean confidence over
the regenerated positions: 
\begin{equation}
C^{(q)}
=
\frac{1}{|\mathcal{M}|}
\sum_{i\in\mathcal{M}}
\gamma_i^{(q)}.
\label{eq:candidate_score}
\end{equation}
For single-arm failures or bimanual failures of both grasps, FoldBack executes the valid candidate with the highest \(C^{(q)}\).

\begin{figure*}[t]
    \centering
    \includegraphics[width=0.95\linewidth]{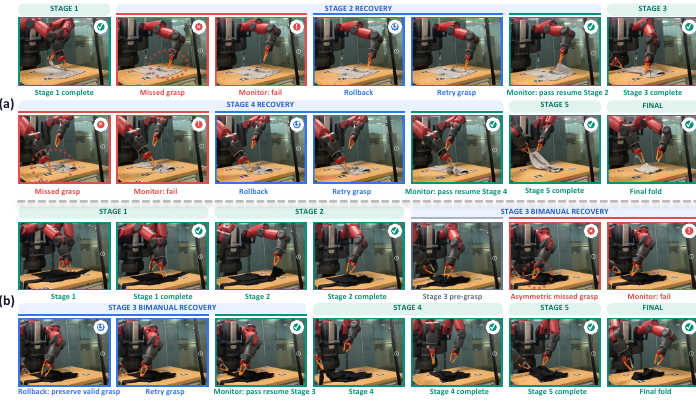}   
    \vspace{-6pt}
    \caption{ Real-world FoldBack recovery.
(a) Short-sleeved T-shirt: recovery from repeated grasp failures during multi-stage folding.
(b) Long-sleeved T-shirt: recovery from an asymmetric bimanual failure while preserving the successful grasp.
    }
    \label{fig:exp}
    \vspace{-3mm}
\end{figure*}

For an asymmetric bimanual failure (Fig. \ref{fig:bimanual}), FoldBack constructs a time-aligned window for each valid resampled motion. 
Each candidate combines the stored motion of the
successful arm, whose gripper remains closed, with the regenerated
motion of the failed arm. Candidates that fail inverse kinematics or violate joint limits are discarded. Each feasible candidate is re-encoded by the frozen tokenizer and evaluated in one batched MGT pass, with \(\mathcal{I}_k\) masked and all other candidate tokens fixed.

Let
\(\bar{\gamma}_i^{(q)}\) denote the probability assigned to the re-encoded
bimanual token at \(i\in\mathcal{I}_k\). The candidate score is
\begin{equation}
\bar{C}^{(q)}
=
\frac{1}{|\mathcal{M}|}
\left(
\sum_{i\in\mathcal{M}\setminus\mathcal{I}_k}
\gamma_i^{(q)}
+
\sum_{i\in\mathcal{I}_k}
\bar{\gamma}_i^{(q)}
\right),
\label{eq:hybrid_candidate_score}
\end{equation}
which replaces the original confidence terms for the failed window with those of the executable bimanual candidate. 
FoldBack executes the feasible candidate with the highest \(\bar{C}^{(q)}\). If no feasible candidate remains, it samples another candidate batch, subject to the same three-batch limit. Once all required grasps are verified after the retry, the re-encoded bimanual subsequence at \(\mathcal{I}_k\) is appended to \(\mathcal{H}^{\mathrm{ver}}\), the corresponding stored confidences are updated, and the remaining trajectory is refined using the new post-lift observation.

\begin{table*}[t]
\vspace{1.5mm}
\centering
\caption{
Real-world long-horizon garment-folding performance. 
`VR', `R\&R', and `SR' denote Verified Rollback,
Rollback-and-Repair, and Structured Repair, respectively. }
\label{tab:main_results}
\scriptsize
\setlength{\tabcolsep}{2.2pt}
\renewcommand{\arraystretch}{1.03}

\begin{tabular*}{\textwidth}{
@{\extracolsep{\fill}}l
cccc cccc cccc@{}
}
\toprule
Method
& \multicolumn{4}{c}{Napkin}
& \multicolumn{4}{c}{Towel}
& \multicolumn{4}{c}{Pants} \\
\cmidrule(lr){2-5}
\cmidrule(lr){6-9}
\cmidrule(lr){10-13}
& SC & RS & FS & IoU
& SC & RS & FS & IoU
& SC & RS & FS & IoU \\
\midrule
DP3
& 71.7 & -- & 56.7 & 0.717
& 60.5 & -- & 53.3 & 0.703
& 66.7 & -- & 60.0 & 0.706 \\
SKIL
& 75.0 & -- & 63.3 & 0.755
& 65.8 & -- & 60.0 & 0.743
& 60.0 & -- & 53.3 & 0.731 \\
MGP-Long
& 78.3 & -- & 70.0 & 0.786
& 73.7 & -- & 66.7 & 0.791
& 73.3 & -- & 66.7 & 0.764 \\
FoldBack w/o R\&R
& 83.3 & -- & 73.3 & 0.828
& 81.6 & -- & 73.3 & 0.815
& 80.0 & -- & 66.7 & 0.786 \\
\midrule
DP3 + VR
& 85.0 & 53.8 & 76.7 & 0.823
& 78.9 & 60.0 & 73.3 & 0.814
& 76.7 & 55.6 & 66.7 & 0.788 \\
FoldBack w/o SR
& 90.0 & 70.0 & 83.3 & 0.853
& 86.8 & 66.7 & 80.0 & 0.846
& 83.3 & 60.0 & 73.3 & 0.826 \\
\textbf{FoldBack(ours)}
& \textbf{93.3} & \textbf{90.9} & \textbf{90.0} & \textbf{0.872}
& \textbf{94.7} & \textbf{85.7} & \textbf{93.3} & \textbf{0.876}
& \textbf{86.7} & \textbf{83.3} & \textbf{80.0} & \textbf{0.851} \\
\bottomrule
\end{tabular*}

\vspace{3pt}

\begin{tabular*}{\textwidth}{
@{\extracolsep{\fill}}l
cccc cccc cccc cccc@{}
}
Method
& \multicolumn{4}{c}{Short-sleeved T-Shirt}
& \multicolumn{4}{c}{Short-sleeved Shirt}
& \multicolumn{4}{c}{Long-sleeved T-Shirt}
& \multicolumn{4}{c}{\textbf{Average}} \\
\cmidrule(lr){2-5}
\cmidrule(lr){6-9}
\cmidrule(lr){10-13}
\cmidrule(lr){14-17}
& SC & RS & FS & IoU
& SC & RS & FS & IoU
& SC & RS & FS & IoU
& SC & RS & FS & IoU \\
\midrule
DP3
& 34.8 & -- & 6.7 & 0.502
& 30.4 & -- & 6.7 & 0.469
& 26.7 & -- & 6.7 & 0.458
& 44.3 & -- & 35.2 & 0.610 \\
SKIL
& 42.0 & -- & 13.3 & 0.558
& 33.3 & -- & 13.3 & 0.501
& 34.7 & -- & 6.7 & 0.548
& 48.7 & -- & 39.0 & 0.656 \\
MGP-Long
& 50.7 & -- & 20.0 & 0.605
& 43.5 & -- & 13.3 & 0.551
& 40.0 & -- & 13.3 & 0.540
& 56.3 & -- & 45.7 & 0.689 \\
FoldBack w/o R\&R
& 63.8 & -- & 33.3 & 0.692
& 56.5 & -- & 26.7 & 0.648
& 54.7 & -- & 26.7 & 0.652
& 67.2 & -- & 53.3 & 0.750 \\
\midrule
DP3 + VR
& 65.2 & 42.1 & 26.7 & 0.674
& 55.1 & 47.1 & 20.0 & 0.627
& 52.0 & 45.0 & 20.0 & 0.619
& 66.3 & 48.9 & 51.4 & 0.738 \\
FoldBack w/o SR
& 75.4 & 61.9 & 46.7 & 0.761
& 66.7 & 55.0 & 33.3 & 0.716
& 66.7 & 63.6 & 40.0 & 0.711
& 76.2 & 61.9 & 62.9 & 0.795 \\
\textbf{FoldBack(ours)}
& \textbf{82.6} & \textbf{79.2} & \textbf{60.0} & \textbf{0.811}
& \textbf{76.8} & \textbf{78.3} & \textbf{53.3} & \textbf{0.782}
& \textbf{77.3} & \textbf{77.8} & \textbf{60.0} & \textbf{0.793}
& \textbf{83.9} & \textbf{80.6} & \textbf{75.2} & \textbf{0.837} \\
\bottomrule
\end{tabular*}
\vspace{-4mm}
\end{table*}

\vspace{-1mm}
\section{Experiments}
\label{sec:experiments}
We evaluate FoldBack in the real world, aiming to answer four questions: (1) \textit{Can it improve the execution of long-horizon garment-folding strategies?} (Sec.~\ref{sec:exp_main}), (2) \textit{Can verified rollback and trajectory plan repair recover from controlled one- and two-corner grasp slips?} (Sec.~\ref{sec:exp_slip}),  (3) \textit{How well does the policy generalize to unseen garments and initial configurations?} (Sec.~\ref{sec:exp_generalization}),  (4) \textit{How do FoldBack's inference-time design choices affect execution and recovery?} (Sec.~\ref{sec:exp_ablation}). All policies are trained on Flat'n'Fold robot demonstrations.

\paragraph{Real-world setup}
We follow the Flat'n'Fold setup which includes a Baxter dual-arm robot and a single ZED2 RGB-D camera. Each point cloud is cropped to the workspace and downsampled to 4,096 points. From the point cloud and robot proprioception, the policy predicts a 16-D action vector as defined in Sec.~\ref{sec:method}. Unless stated otherwise, each trial starts with a flattened garment in the workspace. 
We evaluate on 33 garments from six categories, including napkins of two sizes, three towel instances, pants, short-sleeved T-shirts, long-sleeved T-shirts, and short-sleeved shirts. The garment instances vary in size. The evaluated multi-stage strategies may alternate between single-arm and bimanual manipulation. Some categories use multiple strategies; for towels, the initial orientation determines whether the policy performs a two- or three-stage fold. We train one policy per category jointly on all demonstrated strategies for that category.

\paragraph{Model variants}
We evaluate \textit{FoldBack w/o Rollback-and-Repair}, which retains KAAI but disables grasp-outcome verification, rollback, and trajectory plan repair, and \textit{FoldBack w/o Structured Repair}, which retains KAAI, grasp-verified history, and rollback but replaces structured repair with MGP-Long's~\cite{DBLP:journals/corr/abs-2512-09101} confidence-based remasking of the failed window and remaining future, followed by a single repair candidate.

\paragraph{Baselines}

We compare against the action-chunking policies 3D Diffusion Policy (DP3)~\cite{DBLP:conf/rss/ZeZZHWX24} and SKIL~\cite{DBLP:journals/corr/abs-2501-14400}, and the full-trajectory generative policy MGP-Long~\cite{DBLP:journals/corr/abs-2512-09101}. 
We also evaluate \textit{DP3 + Verified Rollback}, which equips DP3 with the same grasp-outcome monitor and verified rollback as FoldBack without modifying or retraining the base policy.
During normal execution, these baselines perform inference every 12 executed actions, whereas FoldBack refines its trajectory at key-action boundaries. Upon a detected failure, the DP3 variant returns to the pre-grasp configuration, discards the unexecuted actions in the current action chunk, and generates a new action chunk from the resulting observation. DP3, SKIL, and MGP-Long do not explicitly detect grasp failures or initiate retries, and therefore continue execution after a failed grasp.

\paragraph{Evaluation metrics}
A recovery episode starts when a grasp failure is first detected in a folding stage and allows up to three rollback-and-retry attempts; 
further failures during those retries belong to the same episode. The episode ends when the stage completes, and the trial terminates if the stage is still incomplete after the third retry. \emph{Recovery success rate} (RS) is the fraction of episodes whose stage completes within this budget, and \emph{first-retry success rate} (FRS) is the fraction of episodes that complete on the first retry. \emph{Stage completion rate} (SC) is the fraction of required folding stages completed across all trials. We take a reference mask from the final top-down frame of a successful demonstration of the same garment and folding strategy, and we compute the IoU in image coordinates. Final-mask IoU is averaged over all trials, including incomplete ones. A trial counts as a \emph{final folding success} (FS) if every stage completes and the final IoU is at least $0.80$.

\vspace{-1mm}
\subsection{Long-Horizon Garment Folding}
\label{sec:exp_main}

We conduct 30 trials on napkins (15 trials per napkin size) and 15 trials for each other category (i.e. pants, t-shirts, etc.). 

Table~\ref{tab:main_results} shows that MGP-Long is the strongest baseline without recovery. KAAI alone improves average final success from \(45.7\%\) to \(53.3\%\), while full FoldBack reaches \(75.2\%\) final success and \(0.837\) final-mask IoU, achieving the best final success in all six categories; representative recovery executions are shown in Fig.~\ref{fig:exp}. Structured repair further improves final success from \(62.9\%\) to \(75.2\%\). Moreover, the gains are larger on the evaluated five-stage strategies. Against MGP-Long, the stage-completion gains are \(13.4\)--\(15.0\) points on the evaluated two-stage folds and \(31.9\)--\(37.3\) points on the three five-stage garment categories.

\begin{table}[t]
\centering
\caption{Recovery from controlled grasp-slip perturbations on towel folding. `FB', `VR', `SR', and `FF' denote FoldBack, verified rollback, structured repair, and fixed failed window. 
}
\label{tab:controlled_slip}
\small
\setlength{\tabcolsep}{2.0pt}
\renewcommand{\arraystretch}{1}
\begin{tabular}{lccccccc}
\toprule
& \multicolumn{4}{c}{One-Corner Slip }
& \multicolumn{3}{c}{Two-Corner Slip} \\
\cmidrule(lr){2-5}\cmidrule(lr){6-8}
Method
& \makecell{Det.$\uparrow$}
& \makecell{Pres. $\uparrow$}
& \makecell{RS $\uparrow$}
& \makecell{FS $\uparrow$}
& \makecell{Det.$\uparrow$}
& \makecell{RS $\uparrow$}
& \makecell{FS $\uparrow$} \\
\midrule
DP3                            & \NA & \NA & \NA & 6.7 & \NA & \NA & 0.0 \\
DP3 + VR          & 86.7 & 38.5 & 53.8 & 33.3 & 86.7 & 46.2 & 20.0 \\
MGP-Long                       & \NA & \NA & \NA & 13.3 & \NA & \NA & 6.7 \\
FB w/o Rollback & 93.3 & 92.9 & 57.1 & 40.0 & 93.3 & 42.9 & 13.3 \\
FB w/o SR + FF & 93.3 & 64.3 & 50.0 & 26.7 & 100.0 & 40.0 & 20.0 \\
FB w/o SR        & 93.3 & 71.4 & 64.3 & 53.3 & 100.0 & 53.3 & 33.3 \\
\textbf{FoldBack (ours)}        & \textbf{93.3} & \textbf{100.0} & \textbf{85.7} & \textbf{73.3} & \textbf{100.0} & \textbf{80.0} & \textbf{66.7} \\
\bottomrule
\end{tabular}
\vspace{-6mm}
\end{table}

\begin{table*}[t]
\vspace{1.2mm}
\centering
\caption{
Generalization performance, reported as final folding success rate (\%) / final-mask IoU.
S-T, L-T, and S-S denote
short-sleeve T-shirt, long-sleeve T-shirt, and short-sleeve shirt, respectively.
}
\label{tab:generalization}

\scriptsize
\setlength{\tabcolsep}{3.8pt}
\renewcommand{\arraystretch}{1.07}

\begin{tabular}{
l
ccc
@{\hspace{5pt}}
ccccccc
@{\hspace{5pt}}
cc
}
\toprule

& \multicolumn{3}{c}{Initial Orientation (Pants)}
& \multicolumn{7}{c}{Unseen Instance}
& \multicolumn{2}{c}{Non-Canonical} \\

\cmidrule(lr){2-4}
\cmidrule(lr){5-11}
\cmidrule(lr){12-13}

Method
& Seen & Mod. & Large
& Nap. & Tow. & Pants & S-T & L-T & S-S & Avg.
& Nap. & Tow. \\

\midrule

DP3
& 60.0/0.706
& 46.7/0.661
& 33.3/0.603
& 60.0/0.704
& 40.0/0.642
& 50.0/0.671
& 10.0/0.486
& 0.0/0.421
& 10.0/0.458
& 28.3/0.564
& 40.0/0.648
& 33.3/0.612
\\

MGP-Long
& 66.7/0.764
& 60.0/0.724
& 40.0/0.642
& 70.0/0.774
& 50.0/0.710
& 60.0/0.731
& 20.0/0.596
& 10.0/0.521
& 10.0/0.530
& 36.7/0.644
& 53.3/0.704
& 46.7/0.674
\\

\textbf{FoldBack (ours)}
& \textbf{80.0/0.851}
& \textbf{73.3/0.817}
& \textbf{60.0/0.752}
& \textbf{90.0/0.863}
& \textbf{80.0/0.833}
& \textbf{70.0/0.812}
& \textbf{60.0/0.780}
& \textbf{50.0/0.756}
& \textbf{50.0/0.742}
& \textbf{66.7/0.798}
& \textbf{80.0/0.826}
& \textbf{73.3/0.798}
\\

\bottomrule
\end{tabular}
\vspace{-4mm}
\end{table*}

\subsection{Recovery under Controlled Grasp-Slip Perturbations}
\label{sec:exp_slip}

Natural failures vary across trials, so we inject controlled one- or two-corner slips at a predetermined pick event during bimanual towel folding. During lifting, an operator dislodges either one corner (producing one failed and one successful grasp) or both corners (causing both grasps to fail). Each trial contains one perturbation and no subsequent intervention. We conduct 15 trials for each evaluated method--condition pair. \emph{Failure-detection rate} (Det.) is the fraction of injected perturbations detected by the monitor, while \emph{recovery success rate} is the fraction of detected recovery episodes whose affected stage is completed within three recovery attempts. For one-corner slips, \emph{successful-grasp preservation rate} (Pres.) is the fraction of detected asymmetric failures in which the successful grasp remains held from failure detection until recovery or termination. 
To isolate the effect of grasp-verified history, 
we compare FoldBack without Structured Repair with \emph{FoldBack without Structured Repair with a Fixed Failed Window}, which treats the failed window as executed history and keeps it fixed after rollback, refining only the remaining future trajectory.
We also evaluate \emph{FoldBack without Rollback}, which uses the same grasp-outcome monitoring, grasp-verified history, and plan repair but repairs the trajectory directly from the failure configuration without returning to the pre-grasp state. 
We further evaluate the frozen monitor on held-out garments, using its first prediction after each lift to classify the grasp outcome.
Labels are manually annotated independently of the monitor output, and the threshold is selected on the validation set and fixed for testing. Across 96 failed and 352 successful grasps, the monitor achieves \(92.7\%\) failed-grasp recall and an arm-wise false-positive rate of \(4.5\%\). This indicates that the monitor detects most grasp failures while rarely triggering unnecessary recovery after a successful grasp.


Verified rollback improves DP3's final folding success under both perturbations (Table~\ref{tab:controlled_slip}). Under one- and two-corner slips, keeping the failed window editable rather than fixing it after execution raises recovery success by \(14.3\) and \(13.3\) percentage points, respectively, while physical rollback yields corresponding gains of \(28.6\) and \(37.1\) points over FoldBack without Rollback.

Full FoldBack achieves recovery success of \(85.7\%\) and \(80.0\%\) and improves final folding success over FoldBack without Structured Repair by \(20.0\) and \(33.3\) points, respectively. Full FoldBack preserves the successful grasp in \(100\%\) of detected asymmetric failures because it keeps the successful gripper closed and reuses that arm's recorded motion while regenerating only the failed arm. FoldBack without Structured Repair regenerates both arms jointly and preserves the successful grasp in only \(71.4\%\) of detected asymmetric failures.

\vspace{-1mm}
\subsection{Generalization}
\label{sec:exp_generalization}

Table~\ref{tab:generalization} compares DP3, MGP-Long, and FoldBack on unseen initial orientations, held-out garment instances, and non-canonical initial configurations. Qualitative examples are provided in the accompanying video.

\smallskip
\noindent\textbf{Initial-orientation generalization.}
Using pants, we evaluate at the training orientation, where the garment is placed diagonally on the table as in the demonstrations, and two unseen rotation ranges relative to this orientation: moderate (\(5^\circ\!-\!20^\circ\)) and large (\(20^\circ\!-\!30^\circ\)).
Trials at the training orientation reuse the pants trials from Table~\ref{tab:main_results}, while each unseen rotation range contains 15 trials per method. FoldBack achieves final success rates of \(80.0\%\), \(73.3\%\), and \(60.0\%\), compared with \(66.7\%\), \(60.0\%\), and \(40.0\%\) for MGP-Long.

\noindent\textbf{Unseen garment instances.}
For each category, we hold out one physical garment from training and retrain each method on the remaining demonstrations. Ten trials per held-out instance yield 60 trials per method. FoldBack achieves \(66.7\%\) average final success, compared with \(36.7\%\) for MGP-Long and \(28.3\%\) for DP3, and performs best in all six categories.

\noindent\textbf{Non-canonical initial configurations.}
We initialize napkins and towels in near-flat configurations with local wrinkles or a partially folded corner, requiring the robot to flatten before folding.
This evaluates whether the learned flatten-and-fold behavior transfers to unseen non-canonical initial configurations.
Across 15 trials per category, FoldBack achieves \(80.0\%\) and \(73.3\%\) final success on napkins and towels, respectively, versus \(53.3\%\) and \(46.7\%\) for MGP-Long. Examples are provided in the accompanying video.

\begin{table}[t]
\centering
\caption{
Three ablations on towel folding.}
\label{tab:ablation}
\scriptsize
\setlength{\tabcolsep}{0.4pt}
\renewcommand{\arraystretch}{1.08}

\begin{tabular*}{\columnwidth}{
@{\extracolsep{\fill}}
lcc
lcc
lccc
@{}
}
\toprule
\multicolumn{3}{c}{(a) Schedule}
&
\multicolumn{3}{c}{(b) Remasking}
&
\multicolumn{4}{c}{(c) Resampling}
\\
\cmidrule(lr){1-3}
\cmidrule(lr){4-6}
\cmidrule(lr){7-10}
Variant
& SC $\uparrow$
& FS $\uparrow$
& Variant
& RS $\uparrow$
& FS $\uparrow$
& Variant
& FRS $\uparrow$
& RS $\uparrow$
& FS $\uparrow$
\\
\midrule
Fixed-12
& 73.7 & 66.7
& Full future
&75.0&  80.0
& Single
&55.6 & 77.8 & 80.0
\\
Uniform
&76.3 & 66.7
& Failed window
&71.4  &80.0
& Conf.-ranked
&66.7 & 83.3 & 86.7
\\
KAAI
& \textbf{81.6}  & \textbf{73.3}
& Dist.-guided
& \textbf{85.7} & \textbf{93.3}
& Failure-aware
& \textbf{71.4} & \textbf{ 85.7}  & \textbf{93.3}
\\
\bottomrule
\end{tabular*}
\vspace{-5.5mm}
\end{table}

\subsection{Ablation Studies}
\label{sec:exp_ablation}

We conduct three ablations on towel-folding with 15 trials per variant
(Table~\ref{tab:ablation}). To isolate event alignment from update frequency, we compare three refinement strategies using the same trained policy with grasp monitoring and rollback-and-repair disabled. \emph{Fixed-12} refines every 12 executed actions, \emph{Uniform} distributes the same number of refinements as KAAI uniformly over the horizon, and \emph{KAAI} refines before and after each key-action window. KAAI achieves the highest stage completion and final success; because Uniform uses the same number of refinements, this comparison isolates the benefit of event alignment. 
We next compare three remasking strategies. \emph{Full future} regenerates the entire unexecuted trajectory, whereas \emph{Failed window} regenerates only the failed key-action window. \emph{Distance-guided} fully remasks the failed window and progressively lowers the masking ratio over subsequent segments, prioritizing low-confidence tokens. It achieves the highest recovery success (\(85.7\%\)) and final folding success (\(93.3\%\)).
Finally, we compare three candidate-selection strategies. \emph{Single} draws one repair candidate, \emph{Confidence ranking} selects the highest-confidence candidate among \(Q=8\), and \emph{Failure-aware} additionally rejects candidates whose failed-arm pick lies within the exclusion region around previous failed grasps. Failure-aware achieves the highest first-retry, recovery, and final success rates: \(71.4\%\), \(85.7\%\), and \(93.3\%\). Batched resampling takes \(58\,\mathrm{ms}\) per recovery event versus \(37\,\mathrm{ms}\) for a single candidate, adding little inference latency.

\section{Conclusion}

We introduced FoldBack, a self-correcting masked generative framework for long-horizon garment folding. FoldBack maintains a grasp-verified trajectory history so that failed grasp windows can be updated 
rather than becoming fixed trajectory history. Upon failure, it returns the robot to a retryable pre-grasp configuration, selectively repairs the trajectory with distance-guided remasking, and resamples candidates while avoiding previously failed grasp locations. Across six garment categories, FoldBack improves stage completion, recovery success, and final folding performance over action-chunking and full-trajectory baselines. 
Future work will extend FoldBack to more challenging initial configurations that require a separate flattening phase before folding can begin.


\bibliographystyle{IEEEtran}
\bibliography{referencesnew}


\end{document}